\documentclass[sigconf, nonacm]{acmart}
\AtBeginDocument{\providecommand\BibTeX{{\normalfont B\kern-0.5em{\scshape i\kern-0.25em b}\kern-0.8em\TeX}}}
\usepackage{fontawesome5}
\newcommand{\codelink}{https://github.com/EverMind-AI/SkillCorpus/tree/main/skillcorpus/lifecycle}
\newcommand{\datalink}{https://huggingface.co/datasets/EverMind-AI/skillcorpus-lifecycle}

\usepackage{booktabs}
\usepackage{algorithm}
\usepackage[noend]{algpseudocode}
\usepackage{amsmath}

\usepackage{amsmath,amsfonts,bm}

\def\eqref#1{equation~\ref{#1}}

\def\1{\bm{1}}

\DeclareMathAlphabet{\mathsfit}{\encodingdefault}{\sfdefault}{m}{sl}
\SetMathAlphabet{\mathsfit}{bold}{\encodingdefault}{\sfdefault}{bx}{n}

\newtheorem{proposition}{Proposition}

\title[Beyond Gain: Measuring Skill Internalization for Agent Skill Retirement]{Beyond Gain: Measuring Skill Internalization\texorpdfstring{\\}{} for Agent Skill Retirement}

\author{Yanze Wang\textsuperscript{1,2,3,*}, Pengfei Yao\textsuperscript{1,2,*}, Tianyi Sun\textsuperscript{1,2,3},
Hongda Chen\textsuperscript{1,2,4}, Yunyun Han\textsuperscript{1,2},\\
Jun Sun\textsuperscript{3,\dag}, Chuanrui Hu\textsuperscript{1,2,\dag,\ddag}, Yafeng Deng\textsuperscript{1,2,\dag}}
\affiliation{\institution{\textsuperscript{1}EverMind\quad \textsuperscript{2}Shanda Group\quad
\textsuperscript{3}Peking University\quad \textsuperscript{4}Southeast University}\country{}}
\email{\{yanze.wang, yaopengfei, tianyi.sun, hanyunyun, chuanrui.hu, dengyafeng\}@shanda.com\\
hongda.chen@evermind.ai, jsun@pku.edu.cn}

\renewcommand{\shortauthors}{Wang et al.}

\begin{document}

\begin{abstract}
LLM agents retrieve skills from public hubs that hold up to about $10^5$ entries and serve many models.
As models improve, a hub must find the skills a given model has internalized, meaning it can do without
them, and retire them for that model, yet internalization has not been measured with known reliability.
Published rules that compare runs with and without a skill retire on the gain alone. The gain is near
zero both for an internalized skill and when neither the model nor the skill delivers the capability, a
state we call uncovered. We instead withhold the skill, score the model on a task synthesized from it,
and trust a high unaided score only when another model gains from the skill, since nearly half of high
scores come from tasks no model benefits from. Under four models, a published gain rule applied with one
task per skill removes over half of the uncovered skills. On SkillsBench, 28 of its 31 retirements rest
on probes no run passed in full, while retiring on the unaided score costs no detectable reward. Unaided
verdicts reproduce across tasks more often than the gain rule's ($\kappa$ 0.42 against 0.23) but differ
across models, so a hub must read them per model. We release 1,251 skills with tasks and item scores
under four models.

\smallskip
\noindent\raisebox{-0.12em}{\faGithub}\hspace{0.4em}\textbf{Code:} \href{\codelink}{\nolinkurl{EverMind-AI/SkillCorpus}}\\
\raisebox{-0.18em}{\includegraphics[height=1em]{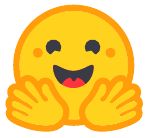}}\hspace{0.4em}\textbf{Data:} \href{\datalink}{\nolinkurl{EverMind-AI/skillcorpus-lifecycle}}
\end{abstract}


\maketitle
\hypersetup{pdfauthor={Yanze Wang, Pengfei Yao, Tianyi Sun, Hongda Chen, Yunyun Han, Jun Sun, Chuanrui Hu, Yafeng Deng}}
{\renewcommand{\thefootnote}{}\footnotetext{\textsuperscript{*}Equal contribution. \textsuperscript{\dag}Corresponding authors. \textsuperscript{\ddag}Project leader.}}

\section{Introduction}
\label{sec:intro}

\begin{figure*}[t]
\centering
\includegraphics[width=\textwidth]{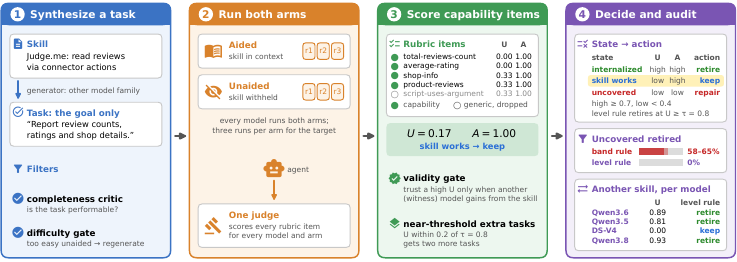}
\caption{The measurement pipeline for one skill. Panel 3 lists the item scores behind $U$ and $A$, with
a dropped generic item in grey. Panel 4 gives this skill's state, the share of uncovered skills that the
band rule $|A-U|<0.10$ and the level rule retire over the four models, and a second skill whose
level-rule verdict differs across models (DS-V4 is DeepSeek-V4-Flash).}
\label{fig:pipeline}
\end{figure*}

Skill libraries are now a Web artefact. Public hubs aggregate agent skills from code repositories and
community registries, the largest holding on the order of $10^5$ entries \citep{skillcorpus}, and agents
retrieve from them at inference time as a search engine retrieves pages. Self-evolving skill libraries that only
accumulate degrade \citep{librarydrift}, and curating a hub raises the questions familiar from Web archives and package registries, what
to keep and what to deprecate on what evidence. The difference is that one entry serves several models whose competence changes between releases.
Like a registry that separates a deprecated package from a broken one, a hub must tell an entry a model has absorbed
from one that does not help. It retires the first for that model while keeping it in the hub, and repairs the second or keeps it for other models. Deciding which skills a model has internalized is therefore an
evaluation problem. The label that drives it comes from agent runs on executable tasks scored by an LLM
judge, and it has to be shown valid and reliable before a hub acts on it.

As models improve, maintainers already remove skill content that looks redundant on benchmark tasks
\citep{saferetire}, and an evaluation framework for skills at scale suggests removing a skill once the
agent solves its task both with and without it and the difference is negligible \citep{tessl}. Neither
turns this practice into a verdict on whole skills, issued for a given model with known reliability. Running the agent
with and without the skill gives two arms, and the gain is the aided score minus the unaided one. The framework's condition
has two parts, a level and a gain, yet the criteria in use keep only the gain or read the aided arm
alone \citep{skillops,librarydrift}. Reinforcement-learning
curricula withdraw a skill during training when the gain falls below zero \citep{skillc} or stops
being positive \citep{skillzero}, and a lifecycle method reads both arms in a diagnostic but retires on the gain
\citep{slim}. Library curation retires an entry when its absolute gain, averaged over development
tasks from the deployment, falls under a threshold \citep{assay}. These criteria serve
their own aims, teaching a model or pruning entries that do not help. A hub must instead judge whether a given
model can do without the skill. Its entries arrive
without tasks and serve users the curator never sees, so a wrong retirement withholds what a user of that model may later
need.

Sorting a library by the outcomes of the two arms gives three states. Where the skill works, the aided
arm passes and the unaided arm fails, and both the gain and the unaided level say keep. Where the model
has absorbed the skill, which we call internalized, both arms pass and both readings say retire. In
the third state, which we call uncovered, the model fails with and without the skill, and the readings
disagree. The unaided level says that the model cannot do
the task, while the gain between two failures is zero, as it is for a skill the model has fully absorbed. Both cases occur in practice. In a paired evaluation of 49 public
software-engineering skills, 39 produced no pass-rate gain \citep{sweskillsbench}. In 24 both arms
passed every task, which the authors attribute to the base model's capability, and in 15 both
passed at the same imperfect rate, which they attribute to deeper capability gaps. The first group
is what a retirement rule is meant to find, and the second looks the same to a gain-only rule.

The measurement literature adds two reasons to doubt the gain. Gain scores are confounded with initial
standing and are less reliable than the scores they are computed from
\citep{cronbach1970change,edwards2001myths}, and value-added research prefers models that condition on
the prior score to models that subtract it \citep{koedel2015vam,mccaffrey2003vam}. The signal is also
rarely measured twice. One curation method bootstraps its decisions
\citep{assay} and a drift study reports per-seed results \citep{librarydrift}, but none of the criteria
\citep{skillc,skillzero,slim,skillops} reports a reliability coefficient.
Retirement should therefore be decided by what the model can do without the skill, not by
how much the skill adds, and that reading needs a known reliability. We call a rule that retires on
the unaided level the level rule.

This paper makes three contributions.

\textbf{Diagnosis.} We formulate retirement as a per-model certificate of unaided capability, and prove that no criterion vanishing at equal arms, including leave-one-out and Shapley
attributions on a single-skill task, can tell an uncovered skill from an internalized one. More runs on a task remove the
noise in this error, not the error itself.

\textbf{Measurement protocol and dataset.} We turn the question into a per-model label for the unaided level, read on
executable tasks that a generator from another model family synthesizes from each skill to state
the goal but not the content the skill supplies. A validity gate trusts a high reading only when another
model gains from the skill, and a reliability decomposition places extra measurement on tasks near the
threshold. We release the task-synthesis pipeline, the protocol and a dataset of 1{,}251 skills with
their tasks, rubrics and item scores under four models.

\textbf{Empirical findings.} Under each of four models, a published rule that retires when the
gain lies in a band around zero, applied with one task per skill, removes more than half of the uncovered
skills. For the target it still removes 45\%, against the level rule's 14\%, when verdict and state are read on different tasks.
On SkillsBench
\citep{skillsbench} almost all of the band's retirements rest on probes that neither arm ever passed in
full, while deleting what the level rule retires costs no detectable reward. The level verdict
reproduces across tasks better than the band's but has to be re-read for each model.

\section{Measuring internalization}
\label{sec:label}

\subsection{Dataset construction}
\label{sec:corpus}

Skills are drawn uniformly at random from a public corpus of agent skills \citep{skillcorpus}. For each
skill, a generator synthesizes one executable task with hidden rubrics in the format of Tessl \citep{tessl}. The task states a goal and may
name the tool or domain, but not the convention the skill teaches or the answer, so a model lacking the
skill must supply that content itself. Two choices make the set adversarial for the target model. First, the
generator comes from a different model family, so the content it embeds is not content the target
trivially shares. Second, a difficulty gate follows generation. The target attempts the task unaided and a
judge scores the attempt. A task already passed on more than half of its rubric is regenerated, and the
skill is dropped if the retry is also too easy. A critic rejects tasks that cannot be performed as
written. Figure~\ref{fig:pipeline} traces one skill, and
Appendix~\ref{app:funnel} gives the counts. As a second library we use Tessl-single, the single-skill
subset of a public skill-evaluation dataset \citep{tessl}, run with the same agent loop, target
and judge (\S\ref{sec:external}).

The target model is Qwen3.6-27B. Qwen3.8-27B, Qwen3.5-397B and DeepSeek-V4-Flash run the same tasks, as
witnesses for the validity gate and as alternative deployment models, and all four are open-weight. The unaided arm is run three
times for every model, and the aided arm three times for the target and twice for each witness, in a
tool-use loop of at most 40 turns. One judge, Qwen3.6-27B, scores every run against the rubrics. Each skill in the released dataset carries one task, and 135 carry a second task generated in
a later batch, which the reliability analysis uses (\S\ref{sec:reliability}).

\subsection{Unaided capability score}
\label{sec:score}

A skill is the library entry as supplied to the agent, the document together with any attached scripts
and references. Its content has three parts, namely the capability it teaches, entry-specific details
such as a project filename that teach nothing, and connective text. A skill is
internalized for a model when the first part is already available to the model. This is a statement
about the model's competence, and its behavioural counterpart is to withhold the skill, score the items
that test the capability, and read the level.

Let $\mathcal{R}$ be the rubric items of a task and $C \subseteq \mathcal{R}$ its capability items
(\S\ref{sec:items}). Running the model $k$ times without the skill gives item outcomes
$y^{0}_{rc} \in \{0, 1\}$ for run $r$ and item $c$, and $k$ runs with the skill give $y^{1}_{rc}$. The
unaided and aided capability scores are
\begin{equation}\label{eq:UA}
U = \frac{1}{k}\sum_{r=1}^{k} \frac{1}{|C|}\sum_{c \in C} y^{0}_{rc}, \qquad
A = \frac{1}{k}\sum_{r=1}^{k} \frac{1}{|C|}\sum_{c \in C} y^{1}_{rc},
\end{equation}
with $k = 3$ for the target model. A gain-based rule reads $A - U$, and the level rule reads $U$ alone.

\subsection{Capability items}
\label{sec:items}

An item is a \emph{capability item} when passing it requires information stated in the skill that a
competent engineer could not reliably supply otherwise. A \emph{generic item} is one any competent
attempt satisfies, such as a file existing or code compiling. An adjudicator,
Qwen3.5-397B, reads the skill and each item and keeps the first kind, 69.2\% of items in the adversarial set and
49.4\% on Tessl-single.

The selection defines what the label measures, and the results do not depend on it. Scoring all task-completion items gives an ordering that correlates with ours at 0.93, and the audit
of \S\ref{sec:gap} reproduces on it, with the band rule of \S\ref{sec:audit} still retiring 51.7\% of the uncovered state and the
level rule still retiring 15.7 points more of the internalized state than the band.

\subsection{Validity gate}
\label{sec:gate}

A high unaided score has two possible causes. Either the model has the capability, or the task does not
require the skill, in which case the score measures task difficulty. One model cannot separate the two.
We therefore admit a task only when another model gains from the skill on the same items. With witness
models $\mathcal{W}$ scored as in Eq.~(\ref{eq:UA}), a task is admitted when
\begin{equation}\label{eq:gate}
\max_{w \in \mathcal{W}} \bigl(A_w - U_w\bigr) > \gamma, \qquad \gamma = 0.1,
\end{equation}
which indicates that the skill carries usable information for the task. One witness that gains is
enough, so evidence accumulates across witnesses and does not expire. Including a weaker model
among them keeps the gate informative as models improve. Two witnesses recover 85\% to 90\% of what three
provide (Appendix~\ref{app:witness}), and the strongest admits the fewest
high-scoring tasks, 101 against 114 and 116.

The gate admits 844 of the 1{,}165 scored tasks, 72.4\%, but only 52.6\% of those with $U \geq 0.8$, so nearly half of the apparent internalization at the top of the range comes from tasks
no other model benefits from. A failing score already rules out the easy-task explanation, so the gate matters
only where the reading is high. A text-only predictor shows the same pattern. It
separates high from low unaided scores at 0.798 AUC ungated and 0.656 gated, so in an ungated
library text mostly tracks whether the task is trivial (Appendix~\ref{app:witness}).

\section{The blind spot of gain-based retirement}
\label{sec:gap}

\subsection{Three states}
\label{sec:states}

Taking 0.7 as a high score and 0.4 as a low one partitions the library into three states, each with
its own correct action. When both $A$ and $U$ are high, the skill is \emph{internalized} and should be
retired. A high $U$ alone shows that the model covers the capability, and a high $A$ only sets aside the
few skills that lower the aided score (Appendix~\ref{app:audit}). When $A$ is high and $U$ is low, the \emph{skill works} and should
be kept. When both are low, the capability is \emph{uncovered}, and the skill should be neither retired
nor relied upon.

\begin{figure}[t]
\centering
\includegraphics[width=\columnwidth]{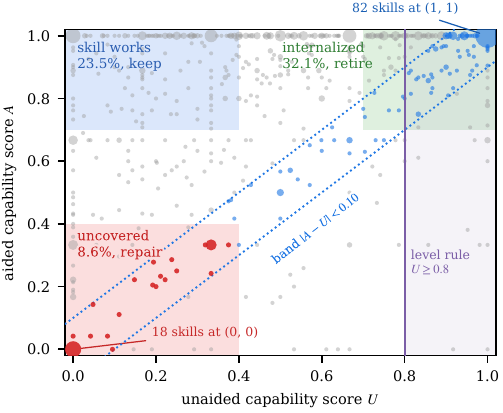}
\caption{The gated adversarial library by unaided and aided score. The band retires the coloured skills between the
dotted lines, the uncovered ones in red, and the level rule retires those right of the
vertical line. Marker area grows with the number of skills at a point.}
\label{fig:states}
\end{figure}

The gain is near zero in the first and third states because the two arms score alike in both. In
Figure~\ref{fig:states} the boundaries of the band rule, which retires when $|A - U| < 0.10$
(\S\ref{sec:audit}), run parallel to $A = U$ and cut across both regions, while
the level rule separates them. The 18 skills at $(0, 0)$ and the 82 at $(1, 1)$ have the same gain, and
the band retires both groups. Retiring a skill in the uncovered state is costly because it records a
capability as absorbed when neither the model nor the skill supplies it. The uncovered state holds
8.6\% of the gated adversarial library and 5.9\% of Tessl-single, with mean scores of 0.10 and 0.15
unaided and 0.18 and 0.27 aided. The band $|A-U|<0.10$ holds 27.8\% and 15.0\% of the two libraries.
Within it the unaided score runs the full range from 0 to 1, with standard deviation 0.331 and 0.277,
so among skills the gain cannot tell apart, capability ranges from none to complete.

\begin{table*}[t]
\centering\small
\caption{What each retirement rule removes, with Qwen3.6-27B as target.}
\label{tab:audit}
\setlength{\tabcolsep}{3.6pt}
\begin{tabular}{lcccc@{\hspace{8pt}}cccc}
\toprule
& \multicolumn{4}{c}{Adversarial} & \multicolumn{4}{c}{Tessl-single} \\
\cmidrule(lr){2-5}\cmidrule(lr){6-9}
Rule & retired & \shortstack{uncovered\\share} & \shortstack{uncovered\\removed} & \shortstack{internalized\\removed}
     & retired & \shortstack{uncovered\\share} & \shortstack{uncovered\\removed} & \shortstack{internalized\\removed} \\
\midrule
gain $< 0$        & 9.4\%  & 11.4\% & 12.3\% & 10.7\% & 2.3\%  & 50.0\% & 19.0\% & 2.7\% \\
gain $\leq 0$     & 25.0\% & 18.5\% & 53.4\% & 45.4\% & 4.8\%  & 35.3\% & 28.6\% & 12.3\% \\
$|$gain$| < 0.10$ & 27.8\% & 17.9\% & 57.5\% & 61.3\% & 15.0\% & 18.9\% & 47.6\% & 45.2\% \\
\midrule
$U \geq 0.8$     & 26.4\% & 0      & 0      & 77.1\% & 13.8\% & 0      & 0      & 67.1\% \\
\bottomrule
\end{tabular}
\end{table*}

\subsection{Formal analysis}
\label{sec:theory}

\paragraph{Setting.} For a skill and a deployment model, let $p_A, p_U \in [0,1]$ be the population
probabilities that the model satisfies the skill's capability items with and without it. With $h = 0.7$
and $l = 0.4$, the three states of \S\ref{sec:states} are the regions
\begin{equation}\label{eq:states}
\begin{gathered}
\mathcal{I} = \{p_A \geq h,\ p_U \geq h\}, \qquad \mathcal{K} = \{p_A \geq h,\ p_U < l\},\\
\mathcal{N} = \{p_A < l,\ p_U < l\}.
\end{gathered}
\end{equation}
Retiring a skill withdraws it from every later task of the deployment model, so the decision has
to establish the unaided capability. The aided score and the gain describe what the skill is worth while it
is present. We therefore treat a retirement verdict as a certificate, issued for one model, that the
model covers the capability, and score a decision $d \in \{\mathrm{retire}, \mathrm{keep}\}$ against a
threshold $\tau$ by
\begin{equation}\label{eq:loss}
\ell(d; p_U) = c_R\, \mathbf{1}[d = \mathrm{retire},\ p_U < \tau] + c_K\, \mathbf{1}[d = \mathrm{keep},\ p_U \geq \tau],
\end{equation}
where $c_R$ is the cost of certifying a capability the model lacks and $c_K$ the cost of withholding the
certificate from one it has absorbed, ordinarily with $c_R \geq c_K$. A retired entry is no longer
retrieved for that model, so an error is not corrected by later use, and a certified uncovered skill
loses the repair flag that the procedure of \S\ref{sec:procedure} raises for it. The loss defines a
correct certificate and is not a measured cost. A rule $g$ maps measurements to a decision, and its risk
on a library whose skills are distributed as $\pi$ is $R(g) = \mathbb{E}_\pi[\ell(g; p_U)]$. We set $\tau = 0.8$ above $h$, so the level rule retires only well inside the internalized region
(Appendix~\ref{app:audit} sweeps $\tau$). The level rule $g_L = \mathbf{1}[p_U \geq \tau]$ attains
$R(g_L) = 0$ by construction. The propositions (Appendix~\ref{app:proofs})
ask what other rules can attain, and \S\ref{sec:audit} measures what they remove.

\begin{proposition}[Contribution criteria cannot see the uncovered state]\label{prop:blind}
Let $\phi(p_A, p_U)$ satisfy $\phi(p, p) = 0$ for every $p$, and let $g$ depend on $(p_A, p_U)$ only
through $\phi$. Then $g$ gives one verdict on the whole diagonal $p_A = p_U$, and
\begin{equation}\label{eq:lb}
R(g) \;\geq\; \min\bigl\{c_R\, \pi(p_A = p_U < \tau),\; c_K\, \pi(p_A = p_U \geq \tau)\bigr\}.
\end{equation}
If $g$ retires when $|\phi| < \varepsilon$, then
$R(g) \geq c_R\, \pi(|\phi| < \varepsilon,\ p_U < \tau)$.
\end{proposition}

\noindent The difference, the log-ratio, the mutual information between arm and outcome, and, for a task
carrying one skill, its leave-one-out or Shapley attribution \citep{ghorbani2019datashapley} all
satisfy the hypothesis. The situation parallels Lord's paradox
\citep{lord1967paradox,holland1982lord}, in which a difference and a level answer different questions,
and retirement asks the question the level answers. On the gated library the band retires 75 skills
whose unaided score is below 0.8, so its plug-in risk is at least $0.089\, c_R$. The two readings agree
wherever the aided arm is at its ceiling, which holds for 75.9\% of skills. There the gain is an affine image of the unaided score,
$A - U \approx 0.954 - U$, and orders the library almost identically up to sign (Appendix~\ref{app:proofs}).
The bound comes from the remaining skills.

\begin{proposition}[Replication does not remove the error]\label{prop:consistency}
With $k$ binary runs per arm, the binary band rule $g_B^{\mathrm{bin}}$ with half-width $\delta$ retires with probability
\begin{equation}\label{eq:pk}
\begin{aligned}
P_k(p_A, p_U) &= \sum_{|a - u| < \delta k} \mathrm{Bin}(a; k, p_A)\, \mathrm{Bin}(u; k, p_U)\\
&\xrightarrow[k \to \infty]{}\; \mathbf{1}\bigl[|p_A - p_U| < \delta\bigr],
\end{aligned}
\end{equation}
so its risk converges to
\begin{equation}\label{eq:rinf}
\begin{aligned}
R_\infty(g_B^{\mathrm{bin}}) ={}& c_R\, \pi\bigl(|p_A - p_U| < \delta,\ p_U < \tau\bigr)\\
&+ c_K\, \pi\bigl(|p_A - p_U| \geq \delta,\ p_U \geq \tau\bigr),
\end{aligned}
\end{equation}
while the level rule applied to the $k$-run mean has $R_k(g_L) \to 0$ whenever $\pi$ puts no mass at
$p_U = \tau$.
\end{proposition}

\noindent Here $p_A$ and $p_U$ are pass probabilities on the given task, so $k$ counts runs of that task.
Across tasks both rules keep the task error of Proposition~\ref{prop:dstudy}, which more tasks reduce and
more runs do not, so the error that replication cannot remove from the band arises within a single task and is not
task noise. At
observed run-level rates, $P_3$ matches the binary column of Table~\ref{tab:targets} to within
about five points. 

\subsection{Auditing published rules}
\label{sec:audit}

\begin{table}[t]
\centering\footnotesize
\caption{Retirement criteria in the literature and in this paper. LOO is leave-one-out, RL is
reinforcement learning, and budget is a cap on retained skills that falls to zero.}
\label{tab:rules}
\setlength{\tabcolsep}{2.1pt}
\begin{tabular}{@{}llll@{}}
\toprule
Method & Setting & Retires when & Guard \\
\midrule
SKILL0 \citep{skillzero}        & RL training & gain $\leq 0$, budget & none \\
SKILLC \citep{skillc}           & RL training & smoothed gain $< 0$, budget & none \\
SLIM \citep{slim}               & RL training & smoothed LOO gain             & exposure, patience \\
ASSAY \citep{assay}             & library     & $|$masked gain$| < 0.10$      & split by spread \\
Library Drift \citep{librarydrift} & library  & aided net success $\leq -0.10$ & 100 trials \\
SkillOps \citep{skillops}       & library     & aided utility $<$ threshold & duplicate exists \\
SkillOS \citep{skillos}         & agent       & learned curator decision     & none stated \\
\midrule
This paper                      & library     & $U \geq 0.8$                 & validity gate, extra tasks \\
\bottomrule
\end{tabular}
\end{table}

Table~\ref{tab:rules} lists the published retirement criteria, and we audit only those whose decision
can be reproduced faithfully on static data. Training-time curricula drop a skill when the gain falls below zero
\citep{skillc} or fails to stay strictly positive \citep{skillzero}, and a budget that falls to zero
withdraws every skill in the end, a curriculum step and not a library edit. We report both training-time rules for reference. A library curation rule, the closer
analogue of our setting, retires when the absolute gain falls under $0.10$ \citep{assay}. The remaining
rules add a guard that draws on evidence gathered while the library is in use, so a bare threshold
would misrepresent them, and we compare them in \S\ref{sec:related} instead. Some of these rules also disclaim the
reading we test, and one states that retirement is not equivalent to successful internalization
\citep{slim}. The audit therefore asks what their verdicts imply once a library acts on them, since a
retired entry no longer supplies its capability.

Write $b^{a}_{r} = \prod_{c \in S} y^{a}_{rc}$ for a binary pass of run $r$ in arm $a$, which
requires full marks on every item of $S$, and $\hat{p}_a = \frac{1}{k}\sum_{r} b^{a}_{r}$ for the pass
rate. On the adversarial set $S = C$, the capability items, and in \S\ref{sec:behav} $S = \mathcal{R}$, every
rubric item, which follows the all-tests-pass success criterion of SkillsBench. The audited rules retire when
\begin{equation}\label{eq:rules}
\begin{aligned}
g_{<0}&:\ A - U < 0, & g_{\leq 0}&:\ A - U \leq 0,\\
g_{B}&:\ |A - U| < \delta, & g_{B}^{\mathrm{bin}}&:\ |\hat{p}_1 - \hat{p}_0| < \delta,
\end{aligned}
\end{equation}
with $\delta = 0.10$, and the level rule $g_L$ retires when $U \geq \tau$ with $\tau = 0.8$. The binary
band $g_B^{\mathrm{bin}}$ uses the binary outcomes the curation rule is defined on, in the single-task
reduction of Appendix~\ref{app:audit}, and $g_B$ applies the same rule to capability scores. If no run in either arm passes every item, then $\hat{p}_1 = \hat{p}_0 = 0$ and
$g_B^{\mathrm{bin}}$ retires the skill whatever its capability scores are, which we call the binary collapse. With $k$ independent runs per
arm this happens with probability
\begin{equation}\label{eq:collapse}
\Pr[\hat{p}_1 = \hat{p}_0 = 0] = (1 - q_1)^k (1 - q_0)^k, \qquad q_a \leq \min_{c \in S} \Pr[y^{a}_{c} = 1],
\end{equation}
where $q_a$ is the chance that a run of arm $a$ passes every item of $S$. A full pass is never more likely
than a pass on the hardest item, so the collapse is more frequent the longer the item set. For each rule,
Table~\ref{tab:audit} gives the share of the library it retires, the share of what it retires that is
uncovered, and how much of the uncovered and of the internalized state it removes.

Retiring on a negative gain is conservative and does little damage. Requiring a strictly positive gain is
not conservative, because a task failed in every run of both arms has a gain of exactly zero. The band
has the same weakness. With binary outcomes the gain is zero whenever the two arms agree on every trial.
That happens both when the task always fails and when it always succeeds, so a criterion centred on
zero cannot tell a task the model always passes unaided from one it never passes even with the skill.
The uncovered state is 2.1 times as dense inside the band as in the library at large, and the band
retires 57.5\% of it, with a skill-level bootstrap interval of $[46.2, 68.7]$. Before retiring, the curation
method rewrites skills whose effect varies widely across tasks into conditional variants. This step cannot run
with one task per skill, and in its native setting it would not reach an uncovered skill, because a task that
always fails or always passes adds nothing to that variation (Appendix~\ref{app:audit}).

The level rule retires nearly the same volume as the band on this target. It cannot retire an
uncovered skill, because it conditions on the unaided level alone, and it retires 15.9 points more of
the internalized state than the band, with a paired interval of $[10.9, 20.7]$. Both comparisons use
states that the unaided score helps define, so they mainly show that the two rules select different
skills. To remove this dependence we read the verdict on one task and the state on the other, in both
directions, over the 124 skills scored on two tasks, which yields 29 decisions on skills uncovered on the other task. The band then retires 45\% of them
and the level rule 14\%, a paired difference of $-31$ points $[-52, -7]$, so the composition of
what the band removes is not an artefact of shared measurement. Restoring the level half of the Tessl condition \citep{tessl}, so that the band retires only when
$A \geq 0.7$, removes no uncovered skill under any of the four targets, and in the cross-fitted audit it retires 7\% of them. It also reproduces across tasks at least as well as
the level rule ($\kappa$ of 0.47 against 0.42 on the two-task skills of \S\ref{sec:reliability}). It does so by reading
a level, since every skill it retires has an unaided score of at least 0.61, and the gain condition that
remains keeps the internalized skills whose two arms differ by 0.1 or more, mostly ones the skill still
lifts (Appendix~\ref{app:audit}). The level rule needs the deployment model's aided arm only to flag
uncovered skills for repair, and it retires those lifted skills as well.

The audit does not depend on which model is the target. Taking each model in
turn as the target, with the other three as witnesses for the gate, gives Table~\ref{tab:targets}. The
dataset was built adversarially against Qwen3.6-27B only, so the other three rows also test whether the
effect survives on tasks not made hard for those models.

\begin{table}[t]
\centering\small
\caption{The audit with each model in turn as target, on 844, 834, 839 and 836 gated skills. Base is
the uncovered base rate.}
\label{tab:targets}
\setlength{\tabcolsep}{3.2pt}
\begin{tabular}{lccccc}
\toprule
& & \multicolumn{2}{c}{Uncovered removed} & \multicolumn{2}{c}{Internalized removed} \\
\cmidrule(lr){3-4}\cmidrule(lr){5-6}
Target & Base & band & binary & band & level \\
\midrule
Qwen3.6-27B  & 8.6\%  & 57.5\% & 79.5\% & 61.3\% & 77.1\% \\
Qwen3.5-397B & 9.4\%  & 62.8\% & 98.7\% & 57.9\% & 72.7\% \\
DeepSeek-V4-Flash  & 15.9\% & 65.4\% & 91.7\% & 62.1\% & 75.3\% \\
Qwen3.8-27B  & 12.8\% & 58.9\% & 86.9\% & 65.9\% & 80.8\% \\
\bottomrule
\end{tabular}
\end{table}

\noindent Under every target the band removes more than half of the uncovered state on capability
scores and 80\% to 99\% on the binary outcomes its source uses. The level rule removes 13
to 16 points more internalized skills than the band.

Two papers argue for the difference. \citet{aces} hold that absolute scores confound skill quality
with agent strength and judge leniency, and \citet{slim} audit on a leave-one-skill-out difference.
For ranking skills they are right, since those confounds cancel in the difference. Retirement asks
whether the capability survives the removal of the skill, and there the agent's strength on the task is
the quantity to be measured. The gain suits ranking skills but not, on its own, retiring them.

\section{Reliability and validity of the label}
\label{sec:worth}

\subsection{Reliability}
\label{sec:reliability}

A label that cannot be reproduced cannot support a retirement decision, and reproducibility is not a
single number. Classical test theory decomposes it into three facets, each corresponding to something
that can be resampled. Internal consistency, over the items of a task, is 0.907. Test--retest
reliability, over runs of one task, is 0.90 at the three-run mean. Parallel-forms reliability,
over a second task for the same skill, is 0.41 (Appendix~\ref{app:reliability}).

Most of the noise therefore comes from which task is drawn, and little from which items are scored or
how many runs are made, so a longer rubric would not help. Proposition~\ref{prop:dstudy} turns this
into a measurement budget (proof in Appendix~\ref{app:proofs}).

\begin{proposition}[Retirement is a mastery decision, and tasks buy more than runs]\label{prop:dstudy}
Let the unaided score over $n_t$ tasks and $n_r$ runs per task have error variance
\begin{equation}\label{eq:sigma}
\sigma_e^2(n_t, n_r) = \frac{\sigma_t^2}{n_t} + \frac{\sigma_r^2}{n_t n_r},
\end{equation}
so that a skill at true level $\mu$ is misclassified with probability approximately
$\Phi\bigl(-|\mu - \tau| / \sigma_e\bigr)$, where $\Phi$ is the standard normal distribution function
and $z_{1-\alpha}$ is its $1-\alpha$ quantile. Then (i) at fixed $n_t$ the error variance cannot fall below
$\sigma_t^2 / n_t$; (ii) if a task costs $c_t$ to generate and a run costs $c_r$, the allocation that
minimizes $\sigma_e^2$ at a fixed budget uses
\begin{equation}\label{eq:nr}
n_r^\ast = \sqrt{\frac{\sigma_r^2\, c_t}{\sigma_t^2\, c_r}}
\end{equation}
runs per task; and (iii) keeping the misclassification probability below $\alpha$ for a skill at
distance $\Delta = |\mu - \tau|$ requires
\begin{equation}\label{eq:nt}
n_t \;\geq\; \frac{z_{1-\alpha}^2\,\bigl(\sigma_t^2 + \sigma_r^2 / n_r\bigr)}{\Delta^2}.
\end{equation}
\end{proposition}

\noindent Classifying a skill against a cut score is the problem of mastery testing, whose adaptive
form rests on sequential tests \citep{wald1945sequential,reckase1983procedure}, and the variance
decomposition is that of generalizability theory \citep{brennan2001gtheory}. Our components are 0.046
for the skill, $\sigma_t^2 = 0.053$ for the task and $\sigma_r^2 = 0.036$ for the run, or 34\%, 39\%
and 27\% of a single run's variance. Eq.~(\ref{eq:nr}) then gives $n_r^\ast = 0.83\sqrt{c_t / c_r}$,
so one run per task is optimal unless generating a task costs more than about three runs. At the same
three runs per skill, three tasks with one run each misclassify 11.4\% of the gated library, one task
run three times misclassifies 16.1\%, and tripling the runs of a single task lowers the error only to
15.2\%. By Eq.~(\ref{eq:nt}) with one run per task and $\alpha = 0.1$, a skill 0.2 from the threshold
needs about four tasks and one 0.3 away at most two, so measurement belongs on tasks near $\tau$.
Adding tasks only within 0.2 of the threshold, 41\% of the gated library, reaches 11.1\% at three quarters the cost of tripling
them everywhere.

Skills at either extreme move by a median of 0.07 when the task is swapped, so one task suffices for
them, while extra tasks for skills in the middle must come from one generation batch. In a pilot that
gave 40 mid-range skills four extra tasks each from one batch, the correlation between two same-batch
pairs of tasks is 0.549, and pairing tasks from different batches lowers it to 0.285.

\begin{figure}[t]
\centering
\includegraphics[width=\columnwidth]{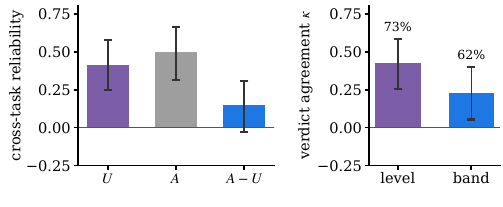}
\caption{Reproducibility across two tasks of a skill, on the 124 skills scored on all task-completion
items of both. Left, parallel-forms reliability. Right, $\kappa$ of each rule's verdict,
with raw agreement above each bar and 95\% bootstrap intervals.}
\label{fig:forms}
\end{figure}

Across runs of one task the aided score and the gain are nearly as reliable as $U$, at 0.85 and 0.80,
and a skill uncovered on two runs reads internalized on the third in 1\% of cases. The 135 skills
that carry a second task, 124 of them scored on both, show how each label survives a change of
task (Figure~\ref{fig:forms}). The gain, at 0.15 across tasks, is far less reproducible than either arm,
and the level rule's verdict agrees with itself more than the band's, 0.42 against 0.23 in Cohen's $\kappa$, with a
paired bootstrap interval of $[0.01, 0.38]$ for the difference. The gain's low reproducibility follows from an identity for
$G = A - U$ measured on two tasks,
\begin{equation}\label{eq:gapcov}
\begin{aligned}
\mathrm{Cov}(G, G') ={}& \mathrm{Cov}(A, A') + \mathrm{Cov}(U, U')\\
&- \mathrm{Cov}(A, U') - \mathrm{Cov}(U, A'),
\end{aligned}
\end{equation}
so whatever a skill carries into both arms on every task cancels. The four terms are 0.044, 0.046,
0.043 and 0.036, so the two arms share 87\% of what persists across tasks and the gain
keeps 0.011.

Reliability also caps what any validation of the label can show. By the attenuation relation, the
correlation of a predictor $X$ with the label satisfies $|\rho_{XU}| \leq \sqrt{\rho_{UU'}}$, where
$\rho_{UU'}$ is the label's reliability. A single-task label therefore caps every downstream validation
at 0.64, and four same-batch tasks lift the cap to 0.84.

\paragraph{Judge dependence.} The target model also scores its own runs, so we re-scored 444 unaided
trajectories with DeepSeek-V4-Flash, the task generator, which is independent of the target though not
of the tasks. The two judges correlate at 0.89, differ by 0.009 in mean, and agree on the retirement
verdict for 91.4\% of trajectories. The disagreements split 20 to 18 between them, so the target model does
not score its own runs systematically higher. Blind to the judge's marks, Claude Opus 5.5, of a third model family, regraded 99 items near the
threshold on the evidence the judge saw. It found 9\% of the judge's decidable full marks unsupported, 15\% counting
partly met criteria, mostly where a number had to be recomputed, and 2 of 20 zero marks wrong. These unsupported full marks
lift $U$ near $\tau$ by about 0.1, the margin between $\tau$ and $h$.

\subsection{Item validity}
\label{sec:itemvalidity}

The validity gate shows that a skill carries usable information for a task, not that each of its
items requires it. The text adjudicator of \S\ref{sec:items} reaches 0.85 precision and 0.81 recall against a
120-item reference annotation by the authors, but text can settle only part of the definition. An item qualifies when the information it tests is stated in the skill,
is not obvious, and is required to pass. The first condition can be decided from text. The last is a
counterfactual, whether the model would pass without the information, and no function of the skill,
task and item identifies it (Appendix~\ref{app:itembehav}). An item requiring that a TypeScript file contain no \texttt{any}
annotation illustrates the problem. The skill states the rule, so the item is kept, yet the model
always satisfies it unaided, as avoiding \texttt{any} is routine.

Such items have no ground truth, but their extent can be bounded. Among the 6{,}802 capability items of gated tasks,
10.0\% are satisfied without the skill by all four models on every run, a conservative estimate of what
survives selection. They concentrate in 29.4\% of skills, at a median of 30\% of an affected skill's
items, and no skill has all of its items in this class. The residue therefore inflates some readings
but leaves no label without diagnostic items.

\subsection{Correcting for the difficulty gate}
\label{sec:truncation}

The difficulty gate of \S\ref{sec:corpus} removed 859 of the 3{,}000 skills attempted, by construction
those most likely to be internalized. When tasks are regenerated without the gate for 115 of them,
69.6\% reach the retirement threshold unaided, against 40.4\% of the 1{,}323 retained skills with an
unaided score. Post-stratifying on $U$ to restore the dropped skills (Appendix~\ref{app:funnel}) lowers
the uncovered base rate from 8.6\% to 7.0\% and its share of what the band retires from 17.9\% to
9.9\%. The share of the uncovered state that the band removes barely moves, to 57.2\%
$[45.7, 68.3]$, because it is conditional on that state.

\begin{algorithm}[t]
\caption{Level-rule retirement of skill $s$ for deployment model $M$}
\label{alg:retire}
\begin{algorithmic}[1]
\Require witnesses $\mathcal{W}$, $\tau = 0.8$, $\gamma = 0.1$, $l = 0.4$, $m = 0.2$
\State $T \gets$ \Call{Synthesize}{$s$}; \ $C \gets$ \Call{CapabilityItems}{$s, T$}
\State $U \gets$ unaided score of $M$ on $C$, three runs \Comment{Eq.~(\ref{eq:UA})}
\If{$|U - \tau| \leq m$}
  \State add two tasks from the same batch, one run each; $U \gets$ mean of the three tasks
\EndIf
\If{$U \geq \tau$}
  \If{$\max_{w \in \mathcal{W}} (A_w - U_w) > \gamma$} \Comment{Eq.~(\ref{eq:gate})}
    \State \Return \textsc{SoftRetire}$(s, M)$
  \ElsIf{$T$ has not been regenerated}
    \State regenerate $T$ and restart \Comment{$T$ does not test $s$}
  \Else
    \State \Return \textsc{Keep}$(s)$ without a verdict
  \EndIf
\EndIf
\If{$U < l$}
  \State $A \gets$ aided score of $M$ on $C$, three runs
  \If{$A < l$} \Comment{uncovered}
    \State \Return \textsc{Keep}$(s)$, flagged for repair
  \EndIf
\EndIf
\State \Return \textsc{Keep}$(s)$
\end{algorithmic}
\end{algorithm}

\subsection{Verdicts across deployment models}
\label{sec:modelbound}

Internalization relates a skill to a model, so we run all four models unaided three times on the same
tasks and apply the same threshold. Between pairs of models, verdicts flip for 14\% to 23\% of skills,
and 37\% of skills receive contradictory verdicts somewhere among the four. Run noise explains little
of this, since resampling a model's three unaided runs flips its verdict for only 5\% to 9\% of skills.
With one task per skill, the skill together with its task explains 62\% of the variance in the unaided
score, the interaction between skill and model 15\%, and overall model strength 0.4\%. Because the
disagreement lies in the interaction and not in
the model main effect, shifting the threshold per model would not remove it. A stronger model's
label cannot stand in for a weaker one's, and a verdict has to be re-read for the model it will govern.
Whichever rule is applied, 19\% to 32\% of what it retires under one model is a skill that works for
another, with $A \geq 0.7$ and $U < 0.4$.

\paragraph{Cost of re-labelling.} A new deployment model reuses the tasks, item selection and witness
evidence and pays only for its own unaided arm, 61\% less than labelling from scratch.

\subsection{From label to decision}
\label{sec:procedure}

Algorithm~\ref{alg:retire} assembles the label into a procedure for one skill and one deployment
model. It measures $U$, adds two tasks when $U$ lies within 0.2 of $\tau$, consults the witnesses only
before retiring, and runs the aided arm only to flag uncovered skills for repair. A skill is internalized only if the model has all of its capability content, but $U$ averages over items.
A lower threshold therefore retires more skills, some of them with content partly missing. This argues for a high threshold, extra measurement near
it and a reversible soft retirement. Appendix~\ref{app:audit} gives the details.

\section{Downstream and external validation}
\label{sec:validation}

We test the label on outcomes outside the audit, first by executing retirement decisions on a public
benchmark and then by repeating the analysis on a second dataset and under automatic skill retrieval.

\subsection{Executing the rules on SkillsBench}
\label{sec:behav}

The audit of \S\ref{sec:audit} shows which skills each rule selects. A curator also needs to know what
a retirement costs, which is measured by what the agent can no longer do once an entry is gone. We therefore make retirement decisions on one measurement, carry them out, and score them
on downstream use they never see.

\paragraph{Retrieval utility.} We first check that retirement leaves retrieval intact. A dense retriever and a reranker rank each
task's own skill among 10{,}000 corpus distractors, and the task is credited with its skill's held-out
gain when the skill ranks in the top three. Retirement is decided on separate runs. The level rule retires 19\% to 29\% of the library across three
models and retains 91\% to 97\% of that utility, against 70\% to 80\% for a random retirement of the same
size.

\paragraph{Benchmark design.} SkillsBench ships each of its tasks with the skills the task is meant to
exercise \citep{skillsbench}. We take all 200 of these skills and build one internalization probe per
skill with the pipeline of \S\ref{sec:label}, which yields 182 scorable probes. Each probe is scored by a
judge that runs code in a sandbox and recomputes numeric and set-valued items before grading, so a claim
made in the transcript does not count as evidence for its own item. The validity gate uses two
witnesses, Qwen3.6-35B-A3B and Qwen3.8-27B, and a probe that it rejects carries no verdict, so its skill
is kept under every rule.

Some probes state the answer to a rubric item in the task text. Those items are dropped before scoring,
but a probe that gives one item away may give away more than the check detects, which would raise $U$.
Before running the benchmark we therefore fixed the analysis to the 92 probes in which no item is given
away, of which the gate admits 59.

Four rules act on the admitted skills. They are the level rule $g_L$, the native binary band
$g_B^{\mathrm{bin}}$ of Eq.~(\ref{eq:rules}), the band $g_B^{\mathrm{all}}$ on the mean over all rubric
items, and a band $g_{B^*}$ that retires the $|D_L|$ admitted skills of smallest $|A - U|$ so that it
matches the level rule in volume. Here $D_g$ is the set of skills rule $g$ retires.

Executing a rule $g$ means deleting every skill in $D_g$ from the environment of each SkillsBench task
that ships it, one slot per such task, and running the task with the remaining skills, under the same agent loop and target
model as the complete library (Algorithm~\ref{alg:bench} in Appendix~\ref{app:behav}). With $\bar r_t$ the mean reward of task $t$
under the complete library and $\bar r_t^{(g)}$ its mean reward after the deletions, the cost of the
rule is
\begin{equation}\label{eq:delta}
\Delta_g = \frac{1}{|\mathcal{T}|}\sum_{t \in \mathcal{T}} \bigl(\bar r_t - \bar r_t^{(g)}\bigr)
\end{equation}
over the 81 benchmark tasks $\mathcal{T}$, where a task that ships none of the retired skills
contributes zero. This matches a library edit, whose cost falls on the tasks that use the retired skill. A positive $\Delta_g$ is reward lost. Two tasks whose reference tests check that the
agent followed the skill's procedure, not that the output is correct, are excluded
(Appendix~\ref{app:behav}). Intervals resample the 81 tasks with 2{,}000 bootstrap draws, and two rules
are compared by the paired difference $\Delta_L - \Delta_g$ on the same draws.

\begin{table}[t]
\centering\small
\caption{Retirements on the 59 admitted probes that give no item away, with the benchmark tasks and
skill slots they touch, the overlap with $g_L$ and the cost $\Delta$ of Eq.~(\ref{eq:delta}) in points of
mean reward. Target Qwen3.6-27B, at least five runs per touched task for each deletion, nine for
the full library.}
\label{tab:behav}
\setlength{\tabcolsep}{3.4pt}
\begin{tabular}{lcccccc}
\toprule
Rule & retired & tasks & slots & shared & $\Delta$ & interval \\
\midrule
$g_L$                 & 18 & 17 & 18 & ---  & $-0.2$ & $[-2.7, 2.4]$ \\
$g_B^{\mathrm{bin}}$  & 31 & 25 & 29 & 10   & $+0.6$ & $[-1.7, 3.1]$ \\
$g_B^{\mathrm{all}}$  & 15 & 14 & 14 & 7    & $-0.5$ & $[-2.0, 0.7]$ \\
$g_{B^*}$             & 18 & 17 & 17 & 9    & $-0.8$ & $[-3.2, 1.7]$ \\
\bottomrule
\end{tabular}
\end{table}

\paragraph{Binary collapse of the native band.} The level rule retires 18 skills and the native band 31,
and they share 10, so the rules remove different skills. A probe records a binary pass only when a trial takes
full marks on every rubric item, and on 28 of the 59 admitted probes no trial in either arm ever did.
For these probes $\hat p_1 = \hat p_0 = 0$, so the band retires them through the binary collapse of
\S\ref{sec:audit}, and they make up 28 of its 31 retirements. Full passes are rare on the admitted probes, 10\% of unaided runs and
37\% of aided ones, the regime in which Eq.~(\ref{eq:collapse}) makes the collapse common. On capability items these probes average
$U = 0.555$, and the skill still raises the score. By the unaided
reading they are not internalized. The band retires them only because the all-items criterion
turns partial success into two zeros.

\paragraph{Benchmark cost.} Deleting what the level rule retires leaves mean reward without detectable change, at
$-0.2$ points $[-2.7, 2.4]$ (Table~\ref{tab:behav}). The level rule touches 17 of the 81 tasks, so per touched
task the 95\% upper limit is about 11 points. No rule shows a cost, while deleting every skill
costs 19.1 points $[11.3, 27.0]$, so the benchmark registers the loss of skills a task still needs. The native band, which retires the most, has
the largest estimate, $+0.6$ points, and the comparison fixed in advance as primary,
$\Delta_L - \Delta_B^{\mathrm{bin}}$, is $-0.8$ points $[-3.1, 1.7]$. The benchmark
therefore detects no difference between the rules, and the choice between them rests on which skills
each removes and on whether its verdict survives a change of task, which the level rule's does more often
than the band's (Figure~\ref{fig:forms}).

\subsection{Replication on Tessl-single}
\label{sec:external}

Tessl-single, which shares an upstream skill collection with the adversarial set, has 849 tasks over
414 skills, and we ran both arms three times. The
same item selection yields 669 scorable tasks over 354 skills, ungated for lack of witness runs. The
structural findings reproduce. Parallel-forms reliability is 0.364 against 0.41, and the uncovered
state is present with a near-zero unaided score. The band holds less of this library (\S\ref{sec:states})
because its tasks depend on their skills more strongly, with a mean gain of 0.384 against 0.306.
Tessl-single is neither gated nor adversarial, so
its shares come closer to those of an ordinary library, and the adversarial set is a stress test.

\subsection{Evidence from automatic skill retrieval}

We also ran the 83 SkillsBench tasks with and without skills retrieved automatically from a
public skill corpus \citep{skillcorpus}, under two harnesses and
two models, Qwen3.6-27B and Qwen3.5-397B. A task passes when its verifier accepts the output. Over the four configurations, 6\% of pairs pass in both arms, 13\% are rescued by the skill, 4\%
are broken by it and 77\% fail in both, so the gain is zero for 83\%, nearly all of them failures in both arms.

\section{Related work}
\label{sec:related}

\paragraph{Curating Web collections.} A skill hub faces the curation problems of other Web collections.
Pages decay and links rot \citep{baryossef2004decay}, package registries let maintainers deprecate
releases \citep{cogo2022deprecation}, and static index pruning removes entries that a retriever rarely
needs \citep{carmel2001pruning}. These settings judge an entry by its own state or retrieval
statistics, whereas a skill's value depends on a model whose competence changes.

\paragraph{Retirement criteria and their guards.} Surveys list removal among the operations that
maintain a skill library \citep{dynamicskills,skillevalsurvey}. Table~\ref{tab:rules} lists the criteria. The rules that
compare arms all read the gain, and the others read the aided arm alone or delegate the decision to a
learned policy. SLIM also labels a skill closer to internalized, as a diagnostic, when its success with
the skill is at least 60\% and disabling it keeps at least 85\% of that success \citep{slim}. This is the
closest prior use of a level. It conditions on the aided level and a ratio between the arms and does not
drive retirement, whereas the level rule thresholds the unaided arm and is the decision itself. A curation method scores utility by leave-one-out replay
\citep{skillbrew}, and others gate admission \citep{grasp,rsea} or govern without removing
\citep{skillsvote}. Follow-up work evicts on the aided net-success statistic of Library Drift \citep{ratchet}, and a judge biased toward false passes silently disables such retirement
\citep{blindcurator}. Redundant clauses of physical-agent skills are retired when utility stays within a margin and
no protected effect appears \citep{saferetire}. To our knowledge none of these rules has been audited for the composition of what it removes, and the closest reports 88\% directional stability of its decisions \citep{assay}.

\paragraph{Fail-both tasks and the case for the difference.} Tasks failing with and without the
skill have been named residual failures and counted per model \citep{regressiontax}, but not tied to a
removal rule. They are excluded where only pairs with a passing arm are admitted
\citep{skillsharmful}, and one evaluation reports 171 cases of zero composite lift without separating
tasks failed in both arms from tasks passed in both, and notes that lift shrinks as models strengthen \citep{aces}.

\paragraph{Reliability, heterogeneity and evaluation.} Reliability reporting ranges from a three-seed
noise floor \citep{singlerewrite} and per-seed results \citep{librarydrift} to single runs
\citep{regressiontax}, and skill effects vary across harnesses, models and setups
\citep{aces,skillsvote,onesizefits,skillcoverage,dynamicskills}. Benchmarks measure skill utility
\citep{skillsbench,skilltester}, and a survey tracks the with--without gain in training \citep{skillevalsurvey}. The critique of the gain follows the gain-score literature
\citep{cronbach1970change,edwards2001myths,koedel2015vam,mccaffrey2003vam} and Lord's paradox
\citep{lord1967paradox,holland1982lord}.

\section{Conclusion}
\label{sec:conclusion}

A skill should be retired when the model can do without it, and the aided--unaided gain cannot tell that
case from one in which neither the model nor the skill delivers the capability. Read on the unaided level with a
validity gate, the verdict separates the two, and executing it on SkillsBench removes skills at no
detectable cost. A hub should therefore retire on an unaided label of known reliability, read for each model it serves.

\bibliographystyle{ACM-Reference-Format}
\bibliography{refs}

\appendix
\section*{Reproducibility statement}
Tasks are synthesized by DeepSeek-V4-Flash, and the completeness critic and the difficulty gate use
Qwen3.6-27B (\S\ref{sec:corpus}). Agents run in a single-agent tool-use loop with shell access to the
task workspace, for at most 40 turns per attempt, and the unaided arm withholds the skill directory and
its text. Every run of every model and arm is scored against the task's rubrics by one judge,
Qwen3.6-27B, and capability items are selected by a text adjudicator, Qwen3.5-397B. The tasks of the 1{,}251
released skills, with their rubrics, capability-item labels, item scores for all four models, the
task-synthesis pipeline with its generator, critic and gate prompts, and the scripts computing the labels
and the rule audits are available on GitHub at \href{\codelink}{\nolinkurl{EverMind-AI/SkillCorpus}},
and the dataset, with a datasheet and per-skill licences, on Hugging Face at
\href{\datalink}{\nolinkurl{EverMind-AI/skillcorpus-lifecycle}}.

\section*{Use of Large Language Models}
Language models are part of the method. They synthesize tasks and rubric items from skill documents,
act as completeness critic and difficulty gate (\S\ref{sec:corpus}), score every run as judge, audit a sample of the judge's marks (\S\ref{sec:reliability}), and
classify rubric items as capability or generic items (\S\ref{sec:items}), whose agreement with a
reference annotation is reported in \S\ref{sec:itemvalidity}. The authors also used a language model to
draft and edit parts of the manuscript. They verified every number, claim and citation against the
experimental records and are responsible for the contents of this paper.

\section{Dataset counts and the difficulty-gate correction}
\label{app:funnel}

\paragraph{Task synthesis.} The generator receives the skill document with its bundled files and returns
one task, its input files and hidden rubrics, of which the label reads the task-completion rubric. It is
instructed to state the goal and the deliverables but not the knowledge that rubric tests, such as an exact
API, parameter value, mapping or required ordering, and to cast the task as a transformation or repair
over the provided inputs, so that a capable agent can reach the goal without the skill but supplies the
skill's specific content only with it. Across the 1{,}251 released tasks the description has a median of
234 words, 78\% come with input files, and the task-completion rubric has a median of 10 items worth 100
points in total, of which the adjudicator keeps a median of 8 as capability items, 69.2\% overall.

\paragraph{Counts.} In Table~\ref{tab:funnel}, the 1{,}251 skills scored in both arms three times are the released dataset
and the 844 that pass the validity gate carry the analysis. Every skill is run in both arms for the
audit, although Algorithm~\ref{alg:retire} needs the aided arm only when $U < l$.
\S\ref{sec:truncation} corrects for the difficulty gate.

\begin{table}[h]
\centering\small
\caption{Dataset construction counts.}
\label{tab:funnel}
\begin{tabular}{lr@{\quad}lr}
\toprule
Attempted & 3{,}000 & Scored unaided once & 1{,}323 \\
\quad too easy & $-859$ & Both arms, three runs & 1{,}251 \\
\quad not performable & $-708$ & Score computable & 1{,}165 \\
\quad generation error & $-34$ & Passes validity gate & \textbf{844} \\
\bottomrule
\end{tabular}
\end{table}

\paragraph{Correcting for the difficulty gate.} We take outcome probabilities within bins of $U$ from
the analysis set, apply them to the 115 regenerated skills, and reweight the 859 dropped skills against
the 1{,}399 that survived generation. Reweighting moves the band's retirements from 27.8\% to 40.3\% of the library,
and \S\ref{sec:truncation} reports the other corrected rates. Restoring the dropped skills adds mass at the
internalized end, which shrinks the uncovered state and enlarges the band. The interval of \S\ref{sec:truncation} resamples both the analysis
set and the 115 regenerated tasks, and the correction assumes that within a bin of $U$ the aided arm
behaves alike in the dropped and retained groups.

\section{Witness models for the validity gate}
\label{app:witness}

The gate admits a task when some witness model gains more than $0.1$ from the skill on the same items.
Two questions bear on its cost, how many witnesses are needed and where in the library.

\paragraph{Where the gate is needed.} Stratifying by the target model's unaided score shows that the
gate is informative only at the top of the range. It admits 52.6\% of the 424 tasks with $U \geq 0.8$,
86.0\% of the 250 between 0.5 and 0.8, 88.8\% of the 249 between 0.2 and 0.5, and 76.4\% of the 242
below 0.2. Witness runs are therefore required only for skills that read high, roughly 36\% of a
library.

\paragraph{How many witnesses.} Taking the union of three witnesses as the reference, single witnesses
recover between 60.7\% and 73.8\% of the tasks the union admits, and pairs recover between 84.8\% and 89.9\%.
The best and worst single witness admit 53.5\% and 43.9\% of tasks, the best and worst pair 65.2\%
and 61.5\%, and all three 72.4\%. A single witness is insufficient because it may carry the same knowledge as the
target, so that both succeed and the task is judged invalid.

\paragraph{A text predictor as an independent check.} A predictor fit to the skill text alone,
evaluated out of fold on the 1{,}165 adversarial skills, separates high from low unaided scores at
0.798 AUC on ungated tasks and at 0.656 once the gate is applied, a difference of $-0.142$ with a
paired bootstrap interval of $[-0.227,-0.050]$. On the 160 skills whose witness evidence is strongest,
those where some witness gains more than 0.5 from the skill, it reaches 0.575 AUC with an interval of
$[0.470, 0.673]$, indistinguishable from chance. Item validity within an admitted task is
the question of \S\ref{sec:itemvalidity}.

\section{Details of the rule audit}
\label{app:audit}

\paragraph{Reduction.} The criteria audited in \S\ref{sec:audit} are all stated on a per-skill score. The
curation rule computes a difference in means over random masks,
$\mathbf{C}[j,i] = \frac{1}{|\mathcal{M}_j^+|}\sum_{k \in \mathcal{M}_j^+} o_k(d_i) - \frac{1}{|\mathcal{M}_j^-|}\sum_{k \in \mathcal{M}_j^-} o_k(d_i)$,
where $o_k(d_i) \in \{0,1\}$ is the outcome of task $d_i$ under mask $k$ and $\mathcal{M}_j^{+}$ and
$\mathcal{M}_j^{-}$ are the masks that include and exclude skill $j$. It retires when the mean over
tasks $\bar{C}(j)$ satisfies $|\bar{C}(j)| < 0.10$. Our tasks carry one skill each with
no co-occurring skills, so this reduces to the aided minus unaided difference on the task, computed
over capability items and averaged across three runs per arm. Skills holding more than one task, which
occur only in Tessl-single, receive the row mean. Before retiring, the method splits each skill
whose spread $H(j) = \max_i \mathbf{C}[j,i] - \min_i \mathbf{C}[j,i]$ reaches $0.40$ into two conditional variants,
because such a skill can have a mean near zero while it helps on some tasks. With one task per skill the spread is zero, so this step never runs (\S\ref{sec:audit}).

\paragraph{Threshold sensitivity.} The boundary separating a high from a low score is not given by
either rule, so we sweep it. Base rate is the share of the library in the uncovered state at that
boundary, enrichment is the ratio of the in-band share to it, and retired is the share of the whole
uncovered state that the band removes (Table~\ref{tab:sweep}).

\begin{table}[t]
\centering\small
\caption{The band audit across the high--low boundary. The last three columns use the gated adversarial
set.}
\label{tab:sweep}
\setlength{\tabcolsep}{2.6pt}
\begin{tabular}{lcccccc}
\toprule
& \multicolumn{3}{c}{Base rate} & & & \\
\cmidrule(lr){2-4}
Bound. & ungated & gated & Tessl & in band & enrich. & retired \\
\midrule
0.3 & 8.7\% & 5.6\%  & 2.8\%  & 13.6\% & 2.45$\times$ & 68.1\% \\
\textbf{0.4} & 11.8\% & \textbf{8.6\%}  & \textbf{5.9\%}  & \textbf{17.9\%} & \textbf{2.07}$\boldsymbol{\times}$ & \textbf{57.5\%} \\
0.5 & 14.1\% & 11.1\% & 7.1\%  & 19.6\% & 1.76$\times$ & 48.9\% \\
0.6 & 17.9\% & 15.4\% & 9.3\%  & 24.3\% & 1.57$\times$ & 43.8\% \\
0.7 & 23.3\% & 21.4\% & 15.8\% & 27.7\% & 1.29$\times$ & 35.9\% \\
\bottomrule
\end{tabular}
\end{table}

The bold row is the main-text setting. Enrichment falls as the boundary loosens, as expected, since a
looser boundary admits less extreme scores. The ordering between datasets is stable, and no setting
removes the effect.

\paragraph{The level rule across its boundary.} The rule of this paper retires when $U \geq \tau$.
Because the uncovered state requires $U < 0.4$, no $\tau \geq 0.4$ can retire any of it, so the
quantities that vary are volume, precision on the internalized state and recall of it. Precision is
below one only where a skill the model passes unaided scores below 0.7 with the skill supplied. Those
entries are not internalized by the two-arm definition, but a skill that lowers performance is not one
to keep either (Table~\ref{tab:tau}).

\begin{table}[t]
\centering\small
\setlength{\tabcolsep}{3pt}
\caption{The level rule across its threshold $\tau$, with the band.}
\label{tab:tau}
\begin{tabular}{lcccccc}
\toprule
& \multicolumn{3}{c}{Adversarial} & \multicolumn{3}{c}{Tessl-single} \\
$\tau$ & retired & precision & recall & retired & precision & recall \\
\midrule
0.7 & 34.7\% & 92.5\% & 100\% & 20.6\% & 100\% & 100\% \\
\textbf{0.8} & \textbf{26.4\%} & \textbf{93.7\%} & \textbf{77.1\%} & \textbf{13.8\%} & \textbf{100\%} & \textbf{67.1\%} \\
0.9 & 18.6\% & 93.6\% & 54.2\% & 5.9\% & 100\% & 28.8\% \\
1.0 & 11.8\% & 94.0\% & 34.7\% & 1.4\% & 100\% & 6.8\% \\
\midrule
band & 27.8\% & 70.6\% & 61.3\% & 15.0\% & 62.3\% & 45.2\% \\
\bottomrule
\end{tabular}
\end{table}

\paragraph{Every state against every rule.} The main text reports three states, and the remaining skills
fall between the thresholds or in the quadrant where the skill lowers the aided score. Each cell counts
the skills of that state that the rule retires, on the gated adversarial set with Qwen3.6-27B as target
(Table~\ref{tab:contingency}). The states are internalized, $A, U \geq 0.7$; skill works, $A \geq 0.7$
and $U < 0.4$; uncovered, $A, U < 0.4$; and skill hurts, $U \geq 0.7$ and $A < 0.4$.

\begin{table}[t]
\centering\small
\setlength{\tabcolsep}{3pt}
\caption{Skills of each state retired by each rule.}
\label{tab:contingency}
\begin{tabular}{lrrrrrr}
\toprule
State & Skills & gain $<0$ & gain $\leq 0$ & band & \shortstack{band,\\$A \geq 0.7$} & level \\
\midrule
Internalized & 271 & 29 & 123 & 166 & 166 & 209 \\
Skill works & 198 & 0 & 0 & 0 & 0 & 0 \\
Uncovered & 73 & 9 & 39 & 42 & 0 & 0 \\
Skill hurts & 7 & 7 & 7 & 0 & 0 & 6 \\
Between thresholds & 295 & 34 & 42 & 27 & 1 & 8 \\
\bottomrule
\end{tabular}
\end{table}

\noindent No rule retires a skill that works. The rules differ on the uncovered state, which only the
level rule and the band restricted to $A \geq 0.7$ leave alone. Of the internalized skills, the level
rule retires 49 that the restricted band keeps, 39 of which the skill still lifts by 0.1 or more and 10
it lowers by as much, and keeps 6 that the restricted band retires, whose unaided score lies between 0.7
and 0.8. The band keeps the few skills that lower the aided score by 0.1 or more.

\section{Proofs of the propositions}
\label{app:proofs}

\paragraph{Proposition~\ref{prop:blind}.} Since $\phi(p, p) = 0$ for every $p$, any $g = g'(\phi)$
takes the single value $g'(0)$ on the diagonal. If $g'(0) = \mathrm{retire}$, every diagonal skill with
$p < \tau$ incurs $c_R$, and otherwise every diagonal skill with $p \geq \tau$ incurs $c_K$. Taking the
smaller of the two gives Eq.~(\ref{eq:lb}). For the band form, every skill with $|\phi| < \varepsilon$
is retired, and those with $p_U < \tau$ each incur $c_R$. The listed measures qualify. The difference and
log-ratio vanish at $p_A = p_U$, the log-ratio taken as zero at $(0, 0)$ as any symmetric smoothing gives. With the arm $X$ uniform on $\{0, 1\}$ and outcome $Y$,
\begin{equation*}
I(X; Y) = H\Bigl(\frac{p_A + p_U}{2}\Bigr) - \frac12\bigl(H(p_A) + H(p_U)\bigr),
\end{equation*}
which is zero exactly when $p_A = p_U$ by strict concavity of the binary entropy $H$. For a task
carrying one skill the leave-one-out and Shapley values of that skill both equal $p_A - p_U$. With other
skills present each marginal contribution is again a difference between two success rates, which
vanishes whenever the task is passed or failed regardless of the skill, but the attribution is then no
longer a function of $(p_A, p_U)$ alone, so the proposition is stated for the single-skill case. The
plug-in figure counts skills retired by $|A - U| < 0.10$ with $U < 0.8$ among the 844 gated skills.

\paragraph{Where the two readings agree.} Where the aided arm saturates, $A$ is approximately a constant
$c$ and the gain is an affine image of the unaided score, $A - U \approx c - U$, so the two order the
library identically up to sign. Within the 75.9\% of the gated adversarial library with $A \geq 0.7$
(\S\ref{sec:theory}), $U$ has standard deviation 0.314 against 0.077 for $A$. The correlation between
$U$ and $A - U$, $-0.67$, is weaker than the $-0.77$ that the shared term alone produces under
independence, so it carries no evidence beyond that shared term.

\paragraph{Proposition~\ref{prop:consistency}.} Let $a \sim \mathrm{Bin}(k, p_A)$ and $u \sim
\mathrm{Bin}(k, p_U)$ be the successes in the two arms. The band retires when $|a/k - u/k| < \delta$,
which gives $P_k$. By the law of large numbers $a/k - u/k \to p_A - p_U$ almost surely, so $P_k \to 1$
when $|p_A - p_U| < \delta$ and $P_k \to 0$ when $|p_A - p_U| > \delta$, and dominated convergence under
$\pi$ gives Eq.~(\ref{eq:rinf}) provided $\pi$ puts no mass on $|p_A - p_U| = \delta$. For the level
rule, $u/k \to p_U$, so the decision converges to $\mathbf{1}[p_U \geq \tau]$ wherever $p_U \neq \tau$,
and its risk to zero. For $k \leq 10$ the resolution $1/k$ is at least $\delta = 0.10$, so the band
retires only when $a = u$ and $P_k = \sum_j \mathrm{Bin}(j; k, p_A)\,\mathrm{Bin}(j; k, p_U)$. This is
why $P_3$ sits slightly above or below its limit instead of at it.

\paragraph{Proposition~\ref{prop:dstudy}.} Under a normal approximation the mean over $n_t$ tasks and
$n_r$ runs is distributed around $\mu$ with variance $\sigma_e^2$ of Eq.~(\ref{eq:sigma}), and the
verdict is wrong when the mean falls on the other side of $\tau$, with probability
$\Phi(-|\mu - \tau| / \sigma_e)$. Part (i) is immediate from Eq.~(\ref{eq:sigma}). For (ii), a
budget $B = n_t (c_t + n_r c_r)$ gives
\begin{equation*}
\begin{aligned}
\sigma_e^2 &= \frac{1}{B}\bigl(\sigma_t^2 + \sigma_r^2 / n_r\bigr)(c_t + n_r c_r)\\
&= \frac{1}{B}\Bigl(\sigma_t^2 c_t + \sigma_r^2 c_r + \sigma_t^2 c_r\, n_r + \frac{\sigma_r^2 c_t}{n_r}\Bigr),
\end{aligned}
\end{equation*}
which is convex in $n_r$ with its minimum at Eq.~(\ref{eq:nr}). For (iii), the
condition $\Phi(-\Delta / \sigma_e) \leq \alpha$ holds exactly when $\sigma_e \leq \Delta / z_{1-\alpha}$.
Substituting Eq.~(\ref{eq:sigma}) and solving for $n_t$ gives Eq.~(\ref{eq:nt}). From the 135 skills that carry a second task
we estimate $\sigma_r^2$ as the variance between the two
runs of the second task, the skill component as the covariance of the two task means after removing the
mean offset between batches, and $\sigma_t^2$ as the remainder of the single-task variance. The
approximation overstates the rate at which verdicts flip between two tasks, 28\% against 25\% observed,
because scores are bounded and concentrated at the extremes, so we use it to compare allocations, not
to predict error rates.

\section{Computing the reliability facets}
\label{app:reliability}

\paragraph{The three states across runs.} When each gated skill is classified from the mean of two
runs and again from the held-out third, 82\% of internalized skills, 72\% of skills in which the skill
works and 71\% of uncovered skills keep their state, and an uncovered skill moves to internalized in 2
of 205 cases, so run noise does not confuse the two states the gain cannot separate. Single-run
agreement over the four classes, the three states and the remainder, is 0.70 with $\kappa = 0.58$.

\paragraph{Computation.} Internal consistency is a split-half correlation over capability items within a
task, corrected by the Spearman--Brown relation, and is an upper bound because items within one task
share a run and a context. Test--retest resamples runs of one task and is reported at the three-run
mean. Parallel forms resamples the task for a fixed skill, using skills that carry more than one task.
The same-batch coefficient of \S\ref{sec:reliability} correlates the means of two tasks with the means of
two others, so by the Spearman--Brown relation it reaches
0.71 with four tasks. The overall coefficient, 0.41, is a one-way intraclass correlation between a
skill's first and second task once the mean offset between the two generation batches is removed, and
0.40 without that adjustment. At three runs the run facet alone caps any predictor's correlation with the
label at 0.95.

\section{Why text cannot establish item validity}
\label{app:itembehav}

\paragraph{Two structural reasons.} First, the items were written from the same texts as the skill and the
task, so an adjudicator reading them shares the generator's information and can check coherence but not
validity. Second, in an agentic task the material an item refers to must be present, or the task could not
be performed, so presence carries almost no variance. Distinctive values from the item appear in the task or its inputs
for 53\% of items, and items whose values appear differ from those whose values do not by 0.074 in pass
rate. Any detector keyed on presence is bounded near the base rate.

\paragraph{The pass rate as a behavioural signal.} The rate at which the target model satisfies an item
is the cheapest behavioural signal for item validity, and it resolves one direction of the problem. An
aggressive adjudicator that removes items it judges weak fails by discarding items the model provably
cannot satisfy, and forbidding it to mark as easy any item the model never satisfies removes that
failure. Our adjudicator is conservative and does not over-remove. Its residual error is keeping items
that are not diagnostic, which the pass rate cannot detect because those are exactly the items the
model passes, so \S\ref{sec:itemvalidity} bounds the residue instead of filtering it.

\section{SkillsBench protocol and excluded tasks}
\label{app:behav}

\begin{algorithm}[h]
\caption{Executing retirement rule $g$ on SkillsBench}
\label{alg:bench}
\begin{algorithmic}[1]
\Require tasks $\mathcal{T}$, each shipping skills $K_t$; retired set $D_g$; complete-library means $\bar r_t$
\For{$t \in \mathcal{T}$}
  \If{$K_t \cap D_g = \emptyset$} $d_t \gets 0$
  \Else
    \State run $t$ with skills $K_t \setminus D_g$; \ $\bar r^{(g)}_t \gets$ mean reward over trials
    \State $d_t \gets \bar r_t - \bar r^{(g)}_t$
  \EndIf
\EndFor
\State $\Delta_g \gets \frac{1}{|\mathcal{T}|}\sum_{t} d_t$ \Comment{Eq.~(\ref{eq:delta})}
\State resample $\mathcal{T}$ 2{,}000 times for the intervals of $\Delta_g$ and $\Delta_L - \Delta_g$
\end{algorithmic}
\end{algorithm}

A retired skill is deleted from each benchmark task environment that ships it. Most skills occupy a
single slot. A few are shipped with two tasks, and a few belong only to tasks outside the 81 scored
ones and occupy none, so the slot counts of Table~\ref{tab:behav} differ from the retired counts and
can exceed the task counts. When two rules leave a task with the same set of skills,
the runs of that condition are pooled and not repeated, and the complete-library baseline pools all
available trials.

Two of the 83 benchmark tasks are excluded from every condition. In \texttt{travel-planning} the
reference test asserts that specific skill-provided functions appear in the tool-call trace, so an
agent that meets 10 of 11 requirements without them scores zero. In \texttt{mario-coin-counting} the
reference answer is derived by sampling one frame per I-frame and the test requires exact equality with
that frame count, so a correct count obtained by sampling once per second is scored wrong. Both tests
reward following the skill's procedure instead of producing the right output, which penalizes every
condition that removes the skill regardless of whether the capability survived. The exclusion was fixed
before any retirement decision and removes the same two tasks from every condition. Keeping the two tasks moves no estimate of Table~\ref{tab:behav} by more than 0.02
points.

Probes that give a rubric item away show why the analysis excludes them. Among admitted probes the level
rule retires 43\% of those that give an item away against 31\% of those that do not, as expected when a
give-away raises $U$. Screening probes for give-aways is therefore part of the protocol.

\end{document}